# Water and Electricity Consumption Forecasting at an Educational Institution using Machine Learning models with Metaheuristic Optimization

Eduardo Luiz Alba [1*], Matheus Henrique Dal Molin Ribeiro [1], Flavio Trojan [1], Gilson Adamczuk Oliveira [1], Erick Oliveira Rodrigues [1]

[1]Graduate Program of Production and Systems Engineering (PPGEPS) Universidade Tecnológica Federal do Paraná, Pato Branco, Brazil.
*eduardoalba0@hotmail.com;

**Abstract**

Educational institutions are essential for economic and social development. Budget cuts in Brazil in recent years have made it difficult to carry out their activities and projects. In the case of expenses with water and electricity, unexpected situations can occur, such as leaks and equipment failures, which make their management challenging. This study proposes a comparison between two machine learning models, Random Forest (RF) and Support Vector Regression (SVR), for water and electricity consumption forecasting at the Federal Institute of Paraná - Campus Palmas, with a 12-month forecasting horizon, as well as evaluating the influence of the application of climatic variables as exogenous features. The data were collected over the past five years, combining details pertaining to invoices with exogenous and endogenous variables. The two models had their hyperparameters optimized using the Genetic Algorithm (GA) to select the individuals with the best fitness to perform the forecasting with and without climatic variables. The absolute percentage errors and root mean squared error were used as performance measures to evaluate the forecasting accuracy. The results suggest that in forecasting water and electricity consumption over a 12-step horizon, the Random Forest model exhibited the most superior performance. The integration of climatic variables often led to diminished forecasting accuracy, resulting in higher errors. Both models still had certain difficulties in predicting water consumption, indicating that new studies with different models or variables are welcome.

# 1 Introduction

In the face of budget cuts in educational institutions in Brazil, it is imperative to explore ways of utilizing resources more efficiently [12]. Within this context, the management of expenses related to water and electricity becomes a challenge, as they can be influenced by various factors such as leaks or equipment overloads. Forecasting consumption can facilitate optimization and aid in the detection of these anomalies.

In forecasting electricity consumption, meteorological variables are commonly employed, given that weather conditions impact the amount of electricity required for heating or cooling environments [19]. Conversely, some studies assert that these climatic variables are not prerequisites for modeling energy consumption [8]. Concerning water consumption, even slight temperature changes or holiday periods can exert a significant influence [7].

Upon analyzing related works, two models have been frequently addressed in the prediction of water and/or electricity consumption. The Random Forest (RF) model has been discussed in studies such as [5, 9, 15–17], while the Support Vector Regression (SVR) model has been employed in works such as [2, 8, 13, 15, 16].

Therefore, the present study conducts a comparative analysis between the machine learning models, Random Forest and Support Vector Regression, for predicting water and electricity consumption in a public educational institution in Brazil. The forecasting horizon considered is 12 months ahead, with hyperparameter optimization achieved using a proposed implementation of a Genetic Algorithm, a stochastic search algorithm inspired by Darwin's theory of evolution [18].

This investigation relies on historical climatic data and monthly consumption data of water and electricity from the Instituto Federal do Paraná (IFPR) - Campus Palmas, in Brazil. The performance of the models is evaluated using measures such as Mean Absolute Percentage Error (MAPE) and Root Mean Squared Error (RMSE).

This study contributes to the literature in several ways. First, by conducting a comparative analysis between Machine Learning models with parameter optimization for predicting water and electricity consumption in an educational institution, it addresses a gap in the literature that lacks this specific theme. Second, the research considers climatic data and historical consumption to enhance the understanding of the variables influencing electricity and water expenditures. Finally, it provides a foundation for comparison that can be of great utility for researchers, professionals, and managers facing similar challenges.

# 2 Materials and Methods

The water and electricity consumption forecast encompasses several stages, including data preprocessing, hyperparameter optimization of machine learning models, model training and testing, and performance evaluation using specific measures, as described through this section.

The data used in the experiment was collected at the Federal Institute of Paraná - Campus Palmas, based on the consumption information from its water and electricity invoices. This dataset includes information for the period from August 2018 to October 2023 for water consumption and from August 2018 to September 2023 for electricity consumption, as demonstrated in Table 1.

**Table 1** Summary of data obtained from water and electricity invoices of the Federal Institute of Paraná - Campus Palmas.

| Consumption | Frequency | Observations | Min. | Max. | Mean | Standard Deviation |
|---|---|---|---|---|---|---|
| Water (mS) | Monthly | 63 | 206 | 1074 | 502.03 | 207.01 |
| Electricity (kWh) | Monthly | 62 | 7252 | 25339 | 15828.60 | 4733.04 |

As shown in Figure 1, an irregular pattern was observed in both time series, particularly during the period from March 2020 to September 2021, when all in-person academic activities were suspended due to the COVID-19 pandemic.

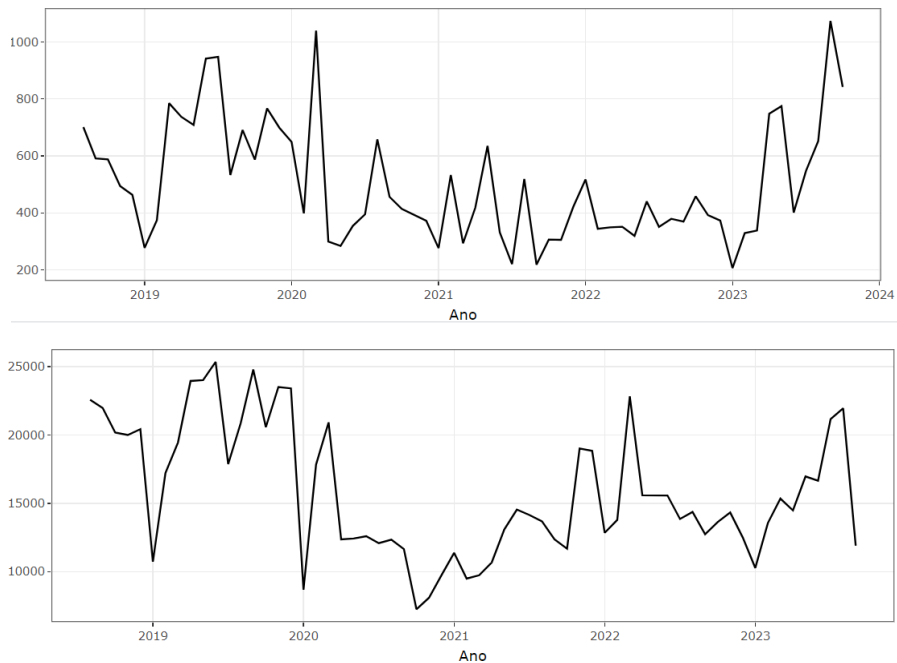

**Fig. 1** Water consumption (in mS) from Aug/2018 to Oct/2023 (top) and electricity consumption (in KWh) from Aug/2018 to Sep/2023 (bottom) at the Federal Institute of Paraná - Campus Palmas.

Due to the decrease in academic activities during school vacation periods, typically between December/January and July/August (according to academic calendars), a seasonal pattern of decreased water and electricity consumption was expected within the time series data. However, the conducted hypothesis tests did not identify any statistically significant trending or seasonal behaviors in the dataset.

To assess the presence of trends in the series, three tests were performed: the Wald-Wolfowitz test [22], the Mann-Kendall test [14], and the Cox-Stuart test [3]. The Wald-Wolfowitz test indicated that the series were not generated randomly, suggesting the presence of trends. However, the Mann-Kendall and Cox-Stuart tests did not identify significant trends in the series, as indicated by the p-values. Finally, the obtained p-valor for Kruskal-Wallis test [10] did not reveal the existence of significant seasonalities. These results are shown in Table 2.

**Table 2** Trend and seasonality hypothesis testing results.

|   | Test | Critical Value | P-Value | Conclusion |
| --- | --- | --- | --- | --- |
| Water | Wald-Wolfowitz (Run Test) | 5% | 0.02117 | Trending |
|  | Man Kendall | 5% | 0.07325 | No trend |
|  | Cox Stuart | 5% | 0.07076 | No trend |
|  | Kruskal-Wallis | 5% | 0.47610 | No seasonality |
| Electricity | Wald-Wolfowitz (Run Test) | 5% | 0.00403 | Trending |
|  | Man Kendall | 5% | 0.09365 | No trend |
|  | Cox Stuart | 5% | 0.47310 | No trend |
|  | Kruskall-Wallis | 5% | 0.47590 | No seasonality |

Furthermore, the tests Augmented Dickey-Fuller (ADF) [4] and Kwiatkowski, Phillips, Schmidt, and Shin (KPSS) [11] were employed to assess whether the mean of the data exhibited stationary behavior. The results of these tests indicated that the time series data did not exhibit stationary properties.

In addition to hypothesis tests used to identify trends, seasonality, and stationarity, Autocorrelation Function (ACF) and Partial Autocorrelation Function (PACF) analyses were also conducted. These analyses indicated that both series exhibited no significant correlations with their past values. This suggests that past values do not hold significant predictive power for forecasting current values.

The exogenous variables (shown in Table 3) incorporated into the Machine Learning Models Random Forest (RF) and Support Vector Regression (SVR) included information related to institutional activities and climatic data. Information regarding the activities of the Federal Institute of Paraná - Campus Palmas was obtained from academic calendars publicly available on its website. Meanwhile, climatic data were acquired through the Agritempo portal, an agrometeorological monitoring system that provides information of various Brazilian municipalities.

**Table 3** Sets of predictor variables for water and electricity consumption.

| Group | Predictor Variables | Type | Frequency |
| --- | --- | --- | --- |
| Institutional Activities | Number of courses in the last hour | Integer | Hourly |
|  | Hours of holiday/break | Integer |  |
|  | Hours of suspended activities (Covid) | Integer |  |
|  | Hours of various activities | Integer |  |
| Weather Data | Minimum Temperature | Decimal | Daily |
|  | Maximum Temperature | Decimal |  |
|  | Average Temperature | Decimal |  |
|  | Precipitation | Decimal |  |
| Time Data | Weekday | Categorical | Daily |
|  | Month | Categorical | Monthly |
| Comsumption Data | Lags | Integer | Monthly |

The proposed methodological framework for this study is illustrated in Figure 2.

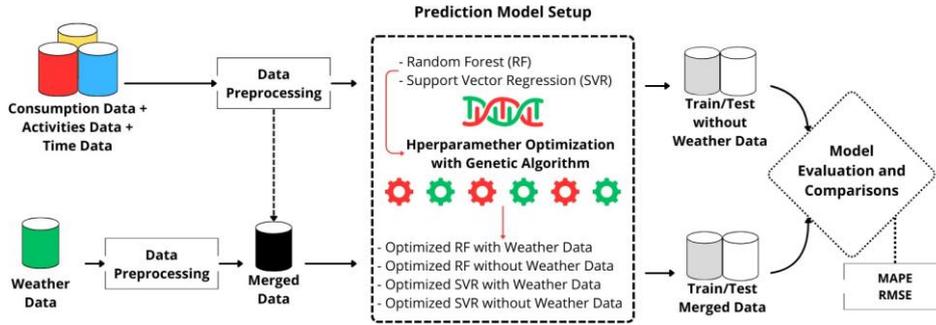

**Fig. 2** Methodology framework proposed for water and electricity consumption forecasting.

Given the disparate frequencies among predictor variables, was imperative preprocessing and transformation procedures to harmonize their temporal resolution with the target variables, which were measured on a monthly basis. This was achieved using Python's grouping functions from the Pandas and NumPy libraries. First, hourly data was grouped into daily data by calculating their averages. Subsequently, the daily values were aggregated into monthly intervals through summation.

Upon conducting a comprehensive review of the literature on water and electricity consumption forecasting, two machine learning models emerged as prominent choices due to their widespread adoption: Random Forest (RF) [5, 9, 15–17] and Support Vector Regression (SVR) [2, 8, 13, 15, 16] exhibited good performances. In light of their wide application, these two models were selected for this investigation.

The Random Forest model is an machine learning algorithm that utilizes multiple decision trees to make predictions. Decision trees are structures that partition data into subsets based on conditional rules. To construct a Random Forest model, numerous decision trees are generated, each trained on a random subset of the data. The final prediction is determined by aggregating the individual predictions of all the trees, typically using the average or majority vote [1].

In the implementation of Random Forest (RF) in the Python scikit-learn library, their hyperparameters include the number of estimators and the maximum depth. The number of estimators directly influences the number of trees employed in the ensemble, while the maximum depth controls complexity level of model. Striking a balance between these parameters is essential, as they significantly impact in accuracy.

The Support Vector Regression (SVR) model is based on the concept of support vectors. Support vectors represent data points that define the boundary separating of them. The SVR seeks to identify a function that minimizes the error between the observed data and the modeled function, with a specified tolerance margin [6, 21].

In the model Support Vector Regression (SVR), the Python scikit-learn library offers various adjustable hyperparameters, among which the kernel, regularization parameter (C), and Epsilon stand out. The kernel determines the function used to map input data, with options such as linear, polynomial, radial, and sigmoid, each being better suited to different types of data and underlying relationships. The regularization parameter controls the adjustment of the model to the training data, avoiding overfitting by balancing complexity and smoothness. The epsilon parameter, width of the margin, defines the acceptable range for prediction errors, influencing the number of support vectors and the trade-off between bias and variance in the model.

Similar to the work of [20], which uses a Grid-Search to determine the optimal values of the hyperparameters for forecast models, this study conducted a custom implementation of the Genetic Algorithm (GA) in Python, considering specific ranges as indicated in Table 4.

**Table 4** Range of values for hyperparameters of RF and SVR models.

|      | Hyperparameter | Data Type | Range of Values |
|------|----------------|-----------|-----------------|
| RF   | estimators     | Integer   | 5 - 200         |
|      | max_depth      | Integer   | 50 - 200        |
| SVR  | Kernel         | String    | "poly", "sigmoid", "rbf" |
|      | Epsilon        | Float     | 0.00001 - 1     |
|      | C              | Float     | 1.0 - 3000.0    |
| Both | lags           | Integer   | 0 - 20          |

The Genetic Algorithm (GA), a powerful metaheuristic optimization technique, draws inspiration from the processes of natural selection and evolution. Its operation commences with the initial creation of a population of individuals, each representing a potential solution to the problem [23].

The initial population is comprised of individuals whose genes represent the parameters and are assigned random values. Each individual within the population possesses a set of genes and a corresponding fitness value. In this instance, fitness is measured using the Mean Squared Error (MSE), which is inversely proportional to the model error. This implies that individuals with lower MSE scores are considered fitter.

In this study, experiments varied the population size (100, 200, and 500) and the number of generations (200, 500, and 1000). Following the initialization of each population, individual fitness was evaluated. Each hyperparamether set were was employed to perform a simple forecast within the dataset. A train-test split validation approach was employed to calculate the Root Mean Squared Error (RMSE) for each individual.

The genetic algorithm execution terminates upon reaching the predetermined number of generations. Within each generation, crossover and mutation operators are applied. Crossover was executed by randomly selecting two individuals and creating a new one by randomly choosing genes from each.

During crossover, there is a probability for the new individual to undergo mutation in one of its genes. The mutation operator randomly modifies the values of a gene, with a difference between 50% and 120% of the original value for numerical genes, or by randomly choosing from the available values for categorical genes.

The training and testing phases of the optimized RF and SVR models involved a train-test split of the datasets (with and without weather data), each containing 12 data for test. The models' performance was then evaluated using two measures: Mean Absolute Percentage Error (MAPE) and Root Mean Squared Error (RMSE).

# 3 Results and Discussions

The results of genetic algorithm optmization on the RF model are shown in Table 5.

**Table 5** Best sets of hyperparameters (individuals) of the Random Forest (RF) model generated by the genetic algorithm in each population and generation variation.

| Forecast | | With climate | | | Without climate | | |
|---|---|---|---|---|---|---|---|
| | Individuals | 100 | 200 | 500 | 100 | 200 | 500 |
| | Generations | 200 | 500 | 1,000 | 200 | 500 | 1,000 |
| Water | Estimators | **6** | 6 | 6 | **6** | 6 | 6 |
| | Max depth | **114** | 52 | 169 | **105** | 50 | 108 |
| | Lags | **4** | 4 | 4 | **4** | 4 | 4 |
| | Time | **89,07** | 209,24 | 461,12 | **88,15** | 203,80 | 454,93 |
| | Fitness | **33.336,08** | 33.336,08 | 33.336,08 | **33.716,00** | 33.716,00 | 33.716,00 |
| Electricity | Estimators | 64 | 64 | **58** | 171 | 7 | **9** |
| | Max depth | 129 | 117 | **137** | 121 | 144 | **131** |
| | Lags | 7 | 7 | **11** | 13 | 13 | **13** |
| | Time | 103,65 | 236,85 | **518,54** | 102,30 | 230,66 | **482,79** |
| | Fitness | 7.566.440,00 | 7.566.440,00 | **7.549.811,75** | 7.525.728,75 | 7.442.614,75 | **7.276.683,33** |

\* Boldfaced values indicate the fittest individuals identified for the corresponding Random Forest experiment.

During the water consumption forecasting RF, while incorporating climatic variables, the average execution time exhibited a increase, with values exceeding 454.93 and 461.12 seconds in the last experiments. Similar observations were noted in the electricity consumption, with average execution times of 518.54 and 482.79 seconds.

In water consumption forecasting experiments, the number of estimators remained constant at 6 for all optimal models, regardless of incorporating climatic variables. However, the maximum depth parameter exhibited a wider range of values with weather data: 52-169 when climatic variables were included and 50-108 without them.

Examining the data of the best-performing individual, the climatic variables in the prediction of electricity consumption, led to an increase in the number of estimators from 9 to 58 for the best-performing individual.

The results of optimization on the SVR model are shown in Table 6.

**Table 6** Best sets of hyperparameters (individuals) of the Support Vector Regression (SVR) model generated by the genetic algorithm in each population and generation variation.

| Forecast | | With climate | | | Without climate | | |
|---|---|---|---|---|---|---|---|
| | Individuals | 100 | 200 | 500 | 100 | 200 | 500 |
| | Generations | 200 | 500 | 1,000 | 200 | 500 | 1,000 |
| Water | Kernel | **poly** | poly | poly | poly | **poly** | poly |
| | Epsilon | **0,2231** | 0,1472 | 0,3024 | 0,4221 | **0,3384** | 0,0351 |
| | C | **3148** | 2659 | 3222 | 791 | **91** | 679 |
| | Lags | **1** | 1 | 1 | 0 | **0** | 0 |
| | Time | **100,60** | 256,03 | 483,62 | 91,84 | **237,12** | 352,19 |
| | Fitness | **43.533,67** | 43.573,33 | 43.566,58 | 46.284,33 | **45.928,33** | 46.158,00 |
| Electricity | Kernel | rbf | rbf | **rbf** | rbf | **rbf** | rbf |
| | Epsilon | 0,6500 | 0,3074 | **0,1805** | 0,0561 | **0,0477** | 0,1410 |
| | C | 2642 | 2483 | **2374** | 2228 | **2256** | 2206 |
| | Lags | 1 | 1 | **1** | 1 | **1** | 1 |
| | Time | 104,48 | 259,86 | **501,53** | 98,07 | **236,33** | 357,75 |
| | Fitness | 9.994.958,08 | 9.915.009,58 | **9.902.730,50** | 9.910.789,83 | **9.909.875,67** | 9.910.789,83 |

\* Boldfaced values indicate the fittest individuals identified for the corresponding SVR experiment.

In water consumption forecasting utilizing Support Vector Regression (SVR), the Genetic Algorithm consistently opted for the polynomial kernel across all scenarios, irrespective of the inclusion of climatic variables. Conversely, the radial basis function (RBF) kernel emerged as the preferred choice for electricity consumption forecasting.

When climatic variables were incorporated into the water consumption forecasting models, the optimal solution entailed the inclusion of only one lag. Conversely, in the absence of these variables, no lag was taken into account. For electricity consumption forecasting, a single lag was universally adopted across all scenarios.

A significant increase in the "C" parameter was observed in water consumption forecasting upon the inclusion of climatic variables, surging from 91 to 3148. These increase was more moderate in electricity consumption, fluctuating from 2374 to 2256.

When forecasting water consumption across a 12-step interval and evaluating performance using Mean Absolute Percentage Error (MAPE) and Root Mean Squared Error (RMSE) measures, the Random Forest (RF) model exhibited superior performance when climatic variables were integrated, as illustrated in Table 7.

**Table 7** Performance of RF and SVR models in water consumption forecast (in $M^3$) at the Federal Institute of Paraná - Campus Palmas.

| Horizon | Measure | RF | | SVR | |
|---|---|---|---|---|---|
| | | With climate | Without climate | With climate | Without climate |
| 12 Months | MAPE (%) | **26,10** | 33,13 | 35,78 | 39,41 |
| | RMSE | **182,58** | 183,62 | 208,65 | 214,31 |

\* Boldfaced values highlight the model with the lowest error for water consumption forecasting.

A visual depiction of water consumption experiment is elaborated in Figure 3.

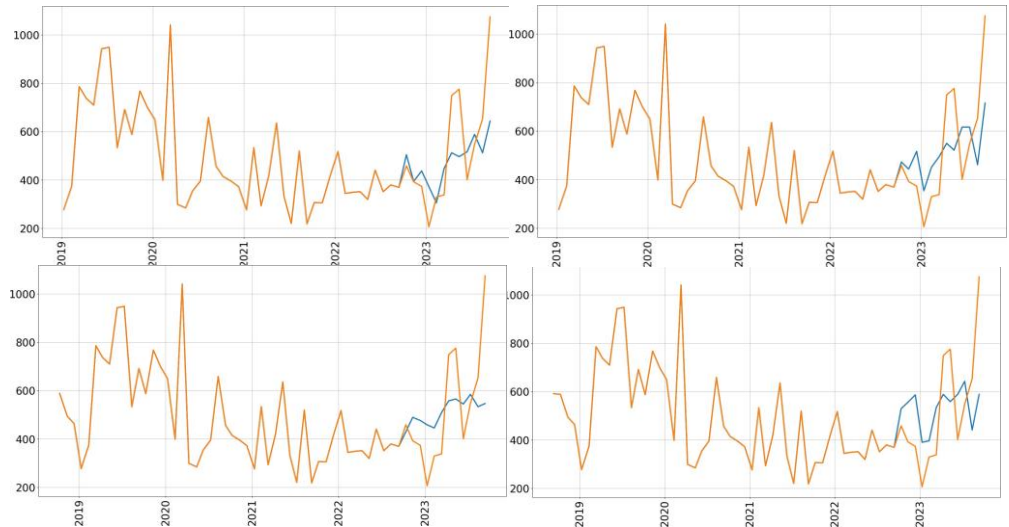

**Fig. 3** Representation of the training and test time series for RF (top) and SVR (bottom) models considering the inclusion of climatic variables (left) and their exclusion (right) for forecasting water consumption 12 steps ahead.

In instances where there exists a tie and the discrepancy in performance between models is minimal, it becomes imperative to take into account the augmented complexity associated with models incorporating climatic variables. Evaluating the difficulty in acquiring forthcoming climatic data becomes essential, as this heightened complexity may introduce forecasting hurdles, potentially resulting in imprecision and the accumulation of errors. Consequently, the selection of the optimal model should encompass considerations of both predictive accuracy and practicality, taking into consideration factors such as data availability.

For 12-step electricity consumption forecasting, the Random Forest (RF) model achieved the lowest Mean Absolute Percentage Error (MAPE) and Root Mean Squared Error (RMSE) when no incorporation of climatic variables was considered, as shown in Table 8.

**Table 8** Performance of RF and SVR models in electricity consumption forecast (in *KWh*) at the Federal Institute of Paraná - Campus Palmas.

| Horizon | Measure | RF | | SVR | |
|---|---|---|---|---|---|
| | | With climate | Without climate | With climate | Without climate |
| 12 Months | MAPE (%) | 12,97 | **12,46** | 15,16 | 15,16 |
| | RMSE | 2.747,69 | **2.697,53** | 3.146,86 | 3.148,00 |

\* Boldfaced values highlight the model with the lowest error for electricity consumption forecasting.

This is further supported by the visual representation in Figure 4.

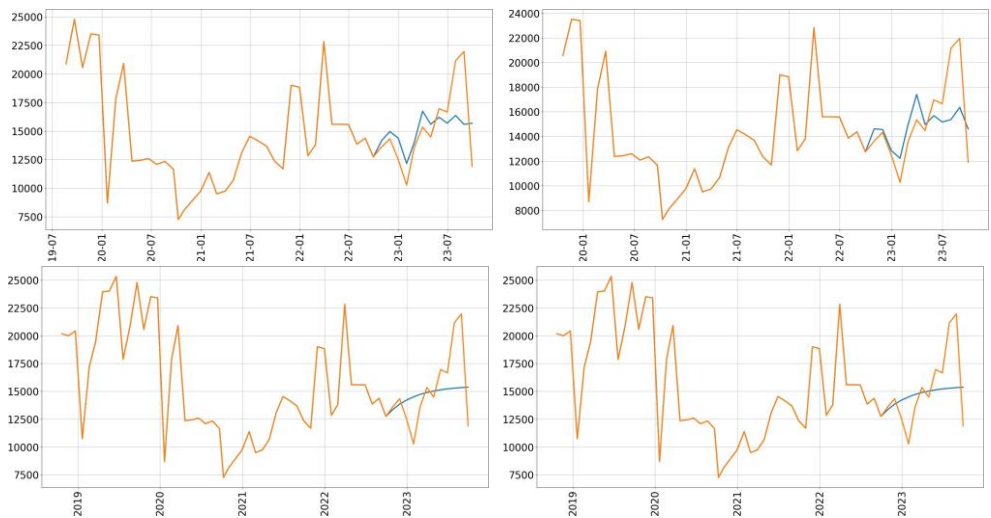

**Fig. 4** Representation of the training and test time series for RF (top) and SVR (bottom) models, with the inclusion of climatic variables (left) and their exclusion (right), for forecasting electricity consumption 12 steps ahead.

The performance of the optimized Random Forest models for predicting water and electricity consumption was compared to that of other methods, as shown in Table 9.

**Table 9** Comparison of the Predictive Performance of Optimized Random Forest models and Alternative Models.

|  | Measure | RF with Climate | Exponential Smoothing | Brown model | Holt-Winters Additive | Holt-Winters Multiplicative |
|---|---|---|---|---|---|---|
| Water | MAPE (%) | **26,1** | 34,62% | 40,38% | 37,41% | 38,23% |
|  | RMSE | **182,58** | 1034,73 | 917,93 | 1077,45 | 951,24 |

|  | Measure | RF without Climate | Exponential Smoothing | Brown model | Holt-Winters Additive | Holt-Winters Multiplicative |
|---|---|---|---|---|---|---|
| Electricity | MAPE (%) | **12,46** | 17,03% | 23,41% | 24,08% | 25,12% |
|  | RMSE | **2697,53** | 12430,95 | 17642,92 | 17578,68 | 17573,09 |

\* Boldfaced values highlight the model with the lowest error for forecasting.

Optimized Random Forest models consistently demonstrated significantly superior performance across all evaluated scenarios. However, traditional methods such as exponential smoothing, Brown, and Holt-Winters, despite being less complex, serve as benchmarks for evaluating the effectiveness of the proposed models.

## 4 Conclusions and Future Work

The genetic algorithm improves the model prediction by optimizing the hyperparameters for the Random Forest and Support Vector Regression models. The goal was to select the hyperparameter combinations that minimized mean squared error.

In a comparative analysis of 12-step ahead forecasts for water and electricity consumption in IFPR, the RF algorithm revealed as the most effective model. In water consumption, the RF model with climate variables outperformed the RF model without climate, achieving a MAPE of 26.10% and a RMSE of 182.58 compared to 33.13% and 183.62. Conversely, in electricity consumption, the RF model without climate achieved a MAPE of 12.46% and RMSE of 2697.53, surpassing the RF model with climate, which had a MAPE of 12.97% and RMSE of 2747.69.

Concerning climatic variables, their impact on forecasting varies depending on several factors, including the forecasting horizon, the specific time series under analysis (whether water or electricity), and the chosen model. In certain instances, their incorporation led to an elevation in errors, suggesting that their utilization should be exercised with prudence. Moreover, obtaining climatic data can pose challenges, potentially introducing a new layer of complexity to the forecasting problem.

In electricity consumption prediction, both models outperformed those applied in water consumption prediction, as indicated by the MAPE. Nevertheless, additional experiments are warranted, encompassing the assessment of alternative time series forecasting models and scrutinizing the impact of additional explanatory variables on the efficacy of both forecasting tasks.

As future work, our objectives include assessing the impact of novel exogenous variables on prediction accuracy and exploring the applicability of machine learning models not covered in this investigation. Additionally, we intend to integrate parallel processing techniques utilizing GPUs to facilitate experiments involving larger numbers of generations within the genetic algorithm framework and better optimization for training. At last, we plan to investigate the influence of data from the COVID-19 pandemic period on the model's performance. This will entail conducting comparative analyses with and without the incorporation of this data.